\documentclass[10pt,twocolumn,letterpaper]{article}

\usepackage{iccv}
\usepackage{times}
\usepackage{epsfig}
\usepackage{graphicx}
\usepackage{amsmath}
\usepackage{amssymb}
\usepackage{threeparttable}
\usepackage{dsfont}
\usepackage{paralist}

\usepackage[breaklinks=true,bookmarks=false]{hyperref}

\iccvfinalcopy % *** Uncomment this line for the final submission

\def\iccvPaperID{3380} % *** Enter the ICCV Paper ID here

\ificcvfinal\fi

\begin{document}

%%%%%%%%% TITLE
\title{ACFNet: Attentional Class Feature Network for Semantic Segmentation}

\author{Fan Zhang$^{1, 2, 3}$\thanks{This work is done when Fan Zhang is an intern at Baidu Inc.} \quad Yanqin Chen$^3$ \quad Zhihang Li$^2$ \quad Zhibin Hong$^{3\dag}$ \\
Jingtuo Liu$^3$ \quad Feifei Ma$^{1, 2}$\thanks{Corresponding Author} \quad Junyu Han$^3$ \quad Errui Ding$^3$ \\ \\
$^{1}$ Laboratory of Parallel Software and Computational Science, \\
Institute of Software, Chinese Academy of Sciences \\
$^{2}$ University of Chinese Academy of Sciences \quad $^{3}$ Baidu Inc. \\
{\tt\small \{zhangf, maff\}@ios.ac.cn} \quad {\tt\small zhihang.li@nlpr.ia.ac.cn}\\
{\tt\small \{chenyanqin, hongzhibin, liujingtuo, hanjunyu, dingerrui\}@baidu.com}
}

\maketitle
\ificcvfinal\thispagestyle{empty}\fi

%%%%%%%%% ABSTRACT
\begin{abstract}
   Recent works have made great progress in semantic segmentation by exploiting richer context, most of which are designed from a spatial perspective. In contrast to previous works, we present the concept of \emph{class center} which extracts the global context from a categorical perspective. This class-level context describes the overall representation of each class in an image. We further propose a novel module, named \emph{Attentional Class Feature} (ACF) module, to calculate and adaptively combine different class centers according to each pixel. Based on the ACF module, we introduce a coarse-to-fine segmentation network, called \emph{Attentional Class Feature Network} (ACFNet), which can be composed of an ACF module and any off-the-shell segmentation network (base network). In this paper, we use two types of base networks to evaluate the effectiveness of ACFNet. We achieve new state-of-the-art performance of 81.85\% mIoU on Cityscapes dataset with only finely annotated data used for training.
\end{abstract}

%%%%%%%%% BODY TEXT
\section{Introduction}

Semantic segmentation, which aims to assign per-pixel class label for a given image, is one of the fundamental tasks in computer vision. It has been widely used in various challenging fields like autonomous driving, scene understanding, human parsing, etc. Recent state-of-the-art semantic segmentation approaches are typically based on convolutional neural networks (CNNs), especially the Fully Convolution Network (FCN) frameworks \cite{long2015fully}.

\begin{figure}[t]
\begin{center}
   \includegraphics[width=1\linewidth]{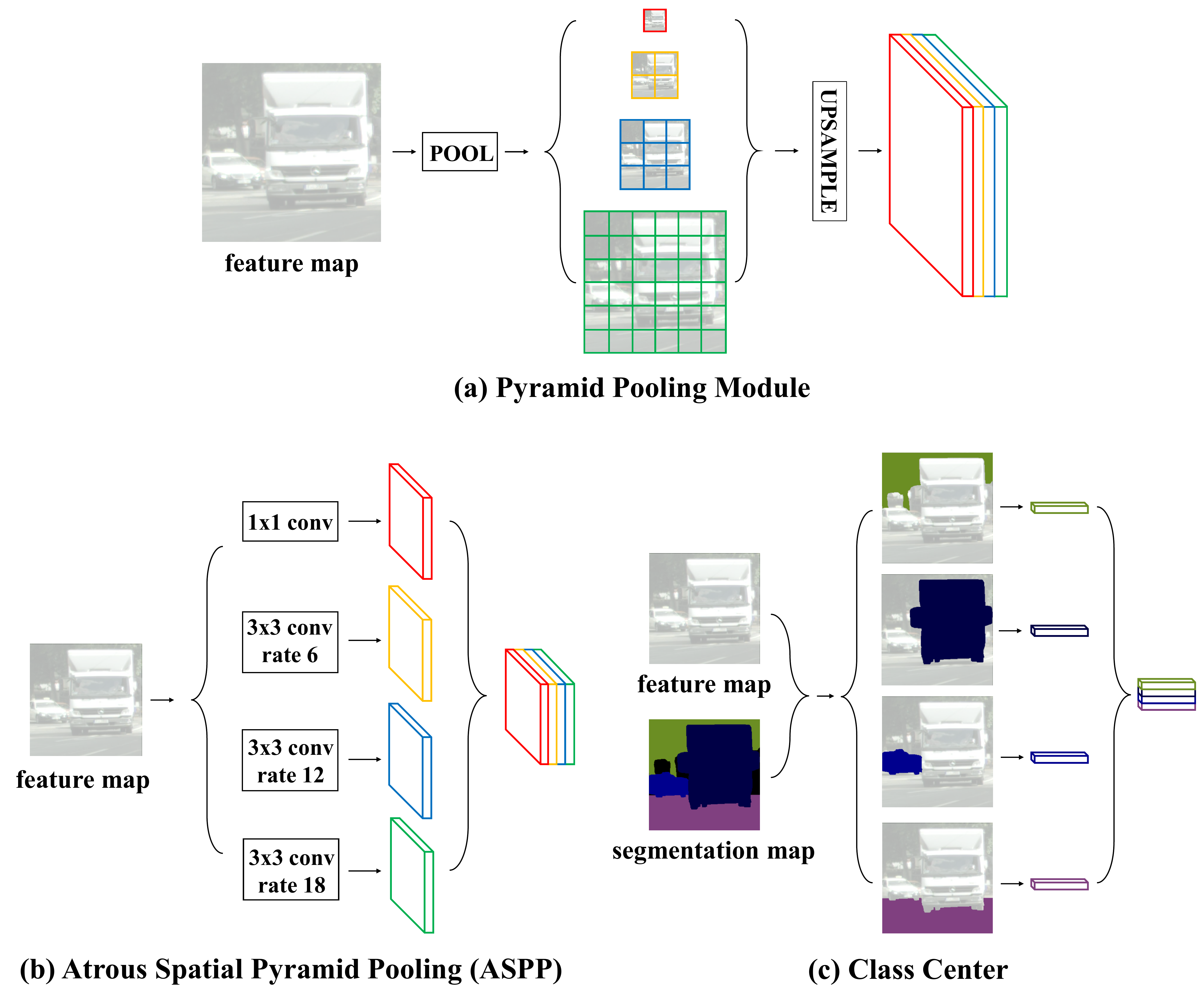}
\end{center}
   \caption{Different approaches to exploit context. The Pyramid Pooling Module (a) and the Atrous Spatial Pyramid Pooling (b) exploit context by employing different spatial sampling strategies. But the Class Center (c) captures the context via a categorical strategy, which uses all pixels of the same category to calculate a class-level feature.}
\label{fig:context} 
\end{figure}

One of the most effective approaches to improve the performance is exploiting richer context \cite{zhao2017pyramid, chen2017rethinking, ding2018context}. For example, Chen \etal \cite{chen2017rethinking} proposed the atrous spatial pyramid pooling (ASPP) to aggregate spatial regularly sampled pixels at different dilated rates around a pixel as its context. In PSPNet \cite{zhao2017pyramid}, the pyramid pooling module divides the feature map into multiple regions with different sizes. The pooled representation of each region is then considered as the context within the same region. Moreover, the global average pooling (GAP) \cite{lin2013network} is also widely used to obtain a global context \cite{yu2018bisenet, zhao2017pyramid, yu2018learning, chen2017rethinking, liu2015parsenet}. Generally, these kinds of methods \cite{chen2018encoder, zhao2017pyramid, ding2018context, yu2018bisenet, yu2018learning} focus on exploiting different spatial strategies to capture richer contextual information. They do not distinguish pixels from different classes explicitly when calculating the context. Surrounding activated objects from different categories contribute the same to the context no matter what category the pixel comes from, which might be confusing for the pixel to determine which category it belongs to.

Different from the methods above, we argue that exploiting the class-level context, an ignored factor before, is also critical for semantic segmentation task. So in this work, we propose a new approach to exploit contextual information from a categorical perspective. We first present a so-called \emph{class center} which describes the overall representation of each category in an image. Specifically, the class center of one class is the aggregation of all features of pixels belonging to this class. A comparison between class center and traditional context modules like ASPP \cite{chen2017rethinking} and pyramid pooling module (PPM) \cite{zhao2017pyramid} is shown in Figure \ref{fig:context}. ASPP and PPM try to exploit context by employing spatial strategies while the class center focuses on capturing the context from a categorical perspective which uses all pixels of the same category to calculate a class-level representation. 

However, it is impractical to get the groundtruth label while testing. Hence, we propose a simple yet effective coarse-to-fine segmentation framework to approximate the class center. The class center for each class can be calculated by the coarse segmentation result and the high-level feature map of the backbone.

Moreover, inspired by the successful applications of attention mechanism in computer vision tasks, \eg \cite{zhao2018psanet, wang2018non, hu2018relation, huang2018ccnet}, we put forward that different pixels need to adaptively pick up to class centers of different categories. For example, if there is no class of `road' in an image, then pixels in this image do not need to focus on feature of `road'. Or if a pixel oscillates between class `person' and class `rider', it should pay more attention to how `person' and `rider' behave in the whole image rather than other categories. Therefore, an \emph{attentional class feature} (ACF) module is proposed to use the attention mechanism to make pixels selectively be aware of different class centers of the whole scene. Different from previous works which design an independent module to learn the attention map, we directly use the coarse segmentation result as our attention map.

\begin{figure*}[t]
\begin{center}
   \includegraphics[width=0.95\linewidth]{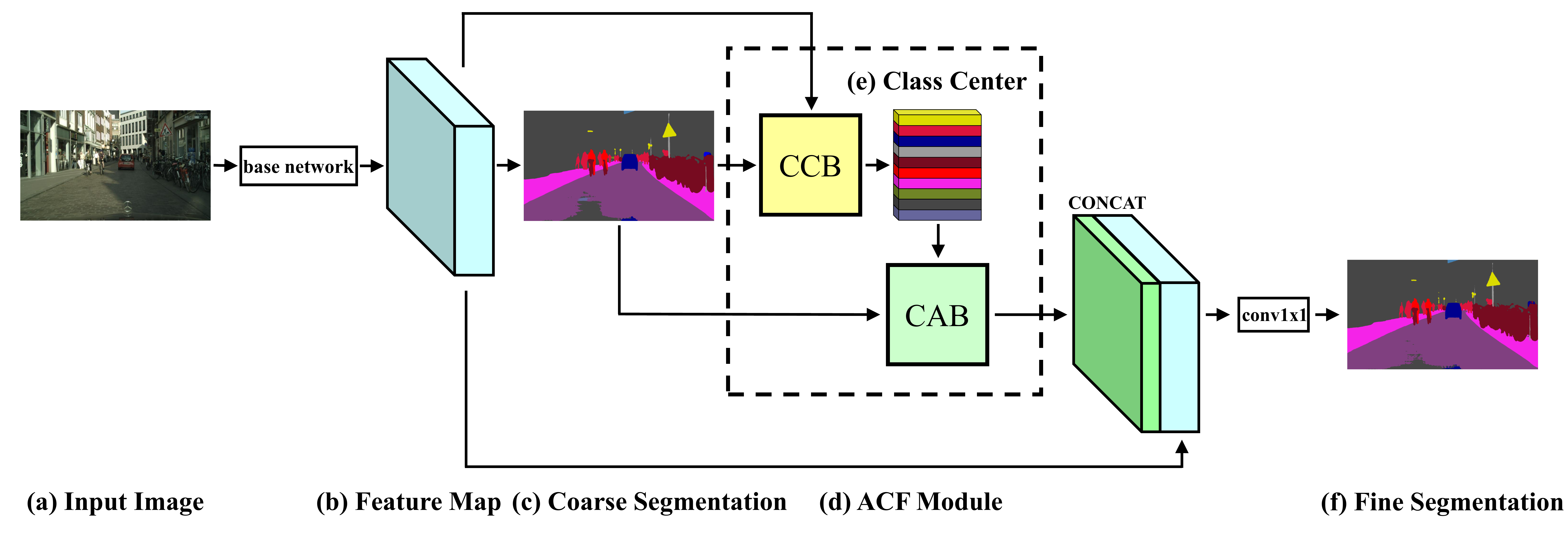}
\end{center}
   \caption{An Overview of the Attentional Class Feature Network. Given an input image (a), we first use a CNN (base network) to get the feature map of the higher layer (b) and the corresponding coarse segmentation result (c). Then an attentional class feature (ACF) module (d) is applied to calculate the class center (e) of different categories and attentional class feature for each pixel according to their coarse segmentation result. Finally the attentional class feature and the feature map (b) are concatenated to get the final fine segmentation (f)}
\label{fig:overall}
\end{figure*}

The overall structure of our proposed coarse-to-fine segmentation network, named \emph{Attentional Class Feature Network}, is shown in Figure \ref{fig:overall}. More specifically, our proposed network consists of two parts. The first part is a complete semantic segmentation network, called \emph{base network}, which generates coarse segmentation results and it can be any state-of-the-art semantic segmentation networks. The second part is our ACF module. The ACF module first uses the coarse segmentation result and the feature map in base network to calculate the class center for each category. After that, the attentional class feature is computed by coarse segmentation result and class center. Finally, the attentional class feature and the original feature in base network are fused to generate the final segmentation.

We evaluate our \emph{Attentional Class Feature Network} (ACFNet) on the popular scene parsing dataset Cityscapes \cite{Cordts2016Cityscapes} and it achieves new state-of-the-art performance of 81.85\% mean IoU with only fine-annotated data for training.

Our contributions can be summarized as follows:

\begin{compactitem}
    \item We first present the concept of \emph{class center}, which represents the class-level context, to help pixels be aware of the performance of different categories in the whole scene.
    \item The \emph{Attentional Class Feature} (ACF) module is proposed to make different pixels adaptively focus on different class centers.
    \item We propose a coarse-to-fine segmentation structure, named \emph{Attentional Class Feature Network} (ACFNet), to exploit class-level context to improve the semantic segmentation.
    \item ACFNet achieves new state-of-the-art performance of the mean IoU of 81.85\% on the popular benchmark Cityscapes \cite{Cordts2016Cityscapes} dataset with only fine-annotated data for training.
\end{compactitem}

%-------------------------------------------------------------------------
\section{Related Work}

\textbf{Semantic Segmentation.} Benefiting from the advances of deep neural networks \cite{krizhevsky2012imagenet, simonyan2014very, szegedy2015going, he2016deep, huang2017densely}, semantic segmentation has achieved great success. The FCN \cite{long2015fully} first replaces the fully connected layer in traditional classification network by convolutional layer to get a segmentation result. SegNet\cite{badrinarayanan2017segnet}, RefineNet \cite{lin2017refinenet}, Deeplabv3+ \cite{chen2018encoder} and UNet \cite{ronneberger2015u} adopt encoder-decoder structure to carefully recover the reduced spatial information through step-by-step upsample operation. Conditional random field (CRF) \cite{chen2014semantic, chandra2017dense, chen2018deeplab}, Markov random field (MRF) \cite{liu2015semantic} and Recurrent Neural Networks (RNNs) \cite{byeon2015scene, shuai2018scene} are also widely used to exploit the long-range dependencies. Dilated convolution \cite{chen2014semantic, yu2015multi} is used to maintain a large enough receptive field while increasing the feature resolution. In our work, we also use the same dilated strategy as in \cite{zhao2017pyramid, chen2017rethinking} to preserve the resolution.

\textbf{Context.} Context plays a critical role in various vision tasks including semantic segmentation. There are bunches of works focusing on how to exploit more discriminative context to help the segmentation. Works like \cite{yu2018bisenet, yu2018learning} use global average pooling (GAP) to exploit the image level context. The atrous spatial pyramid pooling (ASPP) \cite{chen2017rethinking} is proposed to capture the nearby context based on different dilated rate. In PSPNet \cite{zhao2017pyramid}, the average pooling is employed over four different pyramid scales and pixels in one sub-region are treated as the context of pixels within the same sub-region. Some other works focus on how to fuse different context information \cite{yu2018learning, yu2018bisenet, ding2018context, qin2018autofocus} more selectively. In contrast to conventional context described above, in this paper, we harvest the contextual information from a categorical perspective.

More recently, a few works have also investigated the influence of the class-specific context. In EncNet \cite{zhang2018context}, the channel-wise class-level features are enhanced or weakened according to the whole scene. Different from EncNet, we mainly focus on selectively utilizing the class-specific context from the pixel-level in our work.

\textbf{Attention.} Attention is widely used in various fields including natural language processing and computer vision. Vaswani \etal \cite{vaswani2017attention} proposed the transformer using self-attention for machine translation. Hu \etal \cite{hu2018relation} proposed object relation module to extend a learnable NMS operation. The non-local module \cite{wang2018non} is proposed by Wang \etal to calculate the spatial-temporal dependencies. OCNet \cite{yuan2018ocnet} and DANet \cite{fu2018dual} use self-attention mechanism to explore the context. PSANet \cite{zhao2018psanet} also uses an attention map to aggregate long-range contextual information. Our work is inspired by the attention mechanism and we apply it to the calculation of attentional class feature. Instead of designing an independent module to learn the attention map as in previous works, we simply use the coarse segmentation result as the attention map.

\textbf{Coarse-to-fine Methods.} There are a lot of successful applications of using coarse-to-fine approaches, such as face detection \cite{fleuret2001coarse}, shape detection \cite{amit2004coarse}, face alignment \cite{zhu2015face} and optical flow \cite{brox2004high}. Some existing segmentation networks \cite{islam2017label, zhu20173d, wang2011novel, kuo2019shapemask} also adopt coarse-to-fine strategy. Islam \etal \cite{islam2017label} combined high resolution features and coarse segmentation result of low resolution features to get a finer segmentation result. In \cite{zhu20173d}, rough locations of pancreas are obtained in the coarse stage and the fine stage is in charge of smoothing segmentation. In our work, we propose a coarse-to-fine structure and focus on improving the final result through feature-level aggregation.

%------------------------------------------------------------------------
\section{Methodology}

In this section, we first introduce our proposed attentional class feature (ACF) module and elaborate how ACF module captures and adaptively combines the class centers. Then we introduce a coarse-to-fine segmentation structure which consists of our ACF module, named \emph{Attentional Class Feature Network} (ACFNet).

\subsection{Attentional Class Feature Module}

The overall structure of ACF module is shown in Figure.\ref{fig:overall} (d). It consists of of two blocks, Class Center Block (CCB) and Class Attention Block(CAB) which are used to calculate class center and attentional class feature respectively. The ACF module is based on a coarse-to-fine segmentation structure. The input of the ACF module is the coarse segmentation result and the feature map in base network and the output is the attentional class feature.

\subsubsection{Class Center}

The intuition of the concept of \emph{class center} is to exploit richer global context from a categorical view. The class center of class $i$ is defined as the average of features of all pixels belonging to class $i$. Ideally, given the feature map $F \in \mathbb{R}^{C\times H \times W}$, in which $C$, $H$ and $W$ denote the number of channels, height and width of feature map respectively, the class center of class $i$ can be formulated as follows,
\begin{equation}
    F_{class}^i = \frac{\sum_{j=0}^{HW} \mathds{1}[y_j=i] \cdot F_j}{\sum_{j=0}^{HW} \mathds{1}[y_j=i]},
\label{equa:class_feature_ideal}
\end{equation}

where $y_j$ is the label of pixel $j$ and $\mathds{1}[y_j=i]$ is the binary indicator that denotes whether the corresponding pixel comes from the $i$-th class.

Since the groundtruth label is not available during the test phase, we use the coarse segmentation result to evaluate how likely a pixel belongs to a specific class. For a certain class $A$, pixels with higher probability to $A$ in coarse segmentation usually belong to $A$, and these pixels should contribute more when computing the class center of $A$. In this way, we can approximate a robust class center. 

\begin{figure}[t]
\begin{center}
   \includegraphics[width=1\linewidth]{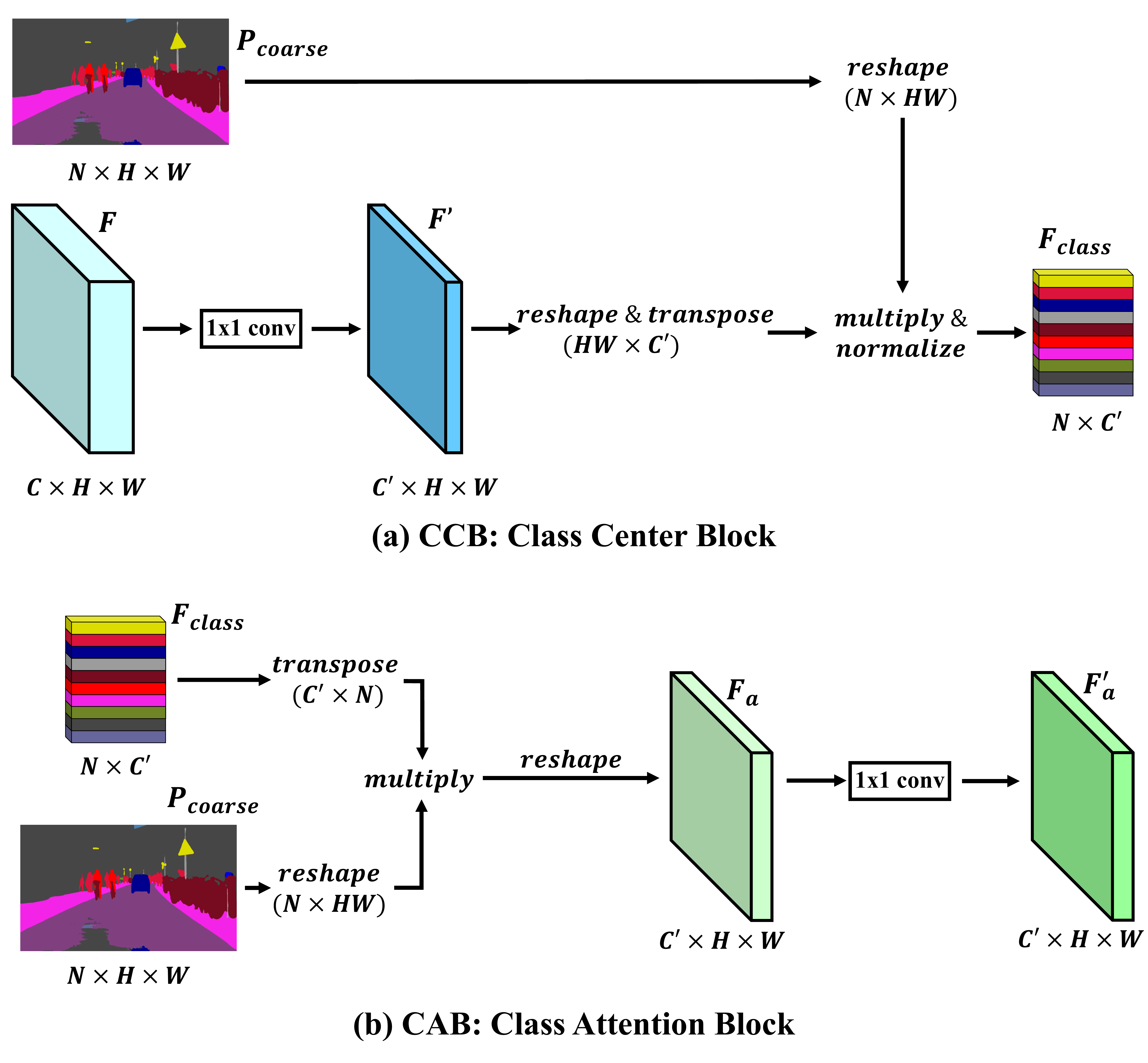}
\end{center}
   \caption{The details of Class Center Block (a) and Class Attention Block (b).}
\label{fig:ccb_acfb}
\end{figure}

Given the coarse segmentation result $P_{coarse} \in \mathbb{R}^{N \times H \times W}$ and the feature map $F \in \mathbb{R}^{C \times H \times W}$, where $N$ is the number of categories, we propose a Class Center Block (CCB) to calculate the class center for each class. The structure of Class Center Block is shown in Figure \ref{fig:ccb_acfb} (a). 

In order to calculate the class center with less computational cost, we first apply a channel reduction operation for feature map through a $1 \times 1$ \emph{conv} to reduce the channel number to $C'$. Then we reshape $P_{coarse}$ to $\mathbb{R}^{N \times HW}$ and the newly calculated feature map $F'$ to $\mathbb{R}^{C' \times HW}$. After that we perform a matrix multiplication and normalization between the $P_{coarse}$ and the transpose of $F'$ to calculate the class centers $F_{class} \in \mathbb{R}^{N \times C'}$. Thus, Equation. \ref{equa:class_feature_ideal} can be rewritten as follows:
\begin{equation}
    F_{class}^i = \frac{\sum_{j=0}^{HW} P_{coarse}^{i,j} \cdot F'_j}{\sum_{j=0}^{HW} P_{coarse}^{i,j}},
\label{equa:class_feature}
\end{equation}

where $P_{coarse}^{i,j}$ denotes the probability of pixel $j$ belonging to class $i$. Both $F'_j$ and $F_{class}^i$ are in $\mathbb{R}^{1 \times C'}$.

\begin{figure}[t]
\begin{center}
   \includegraphics[width=1\linewidth]{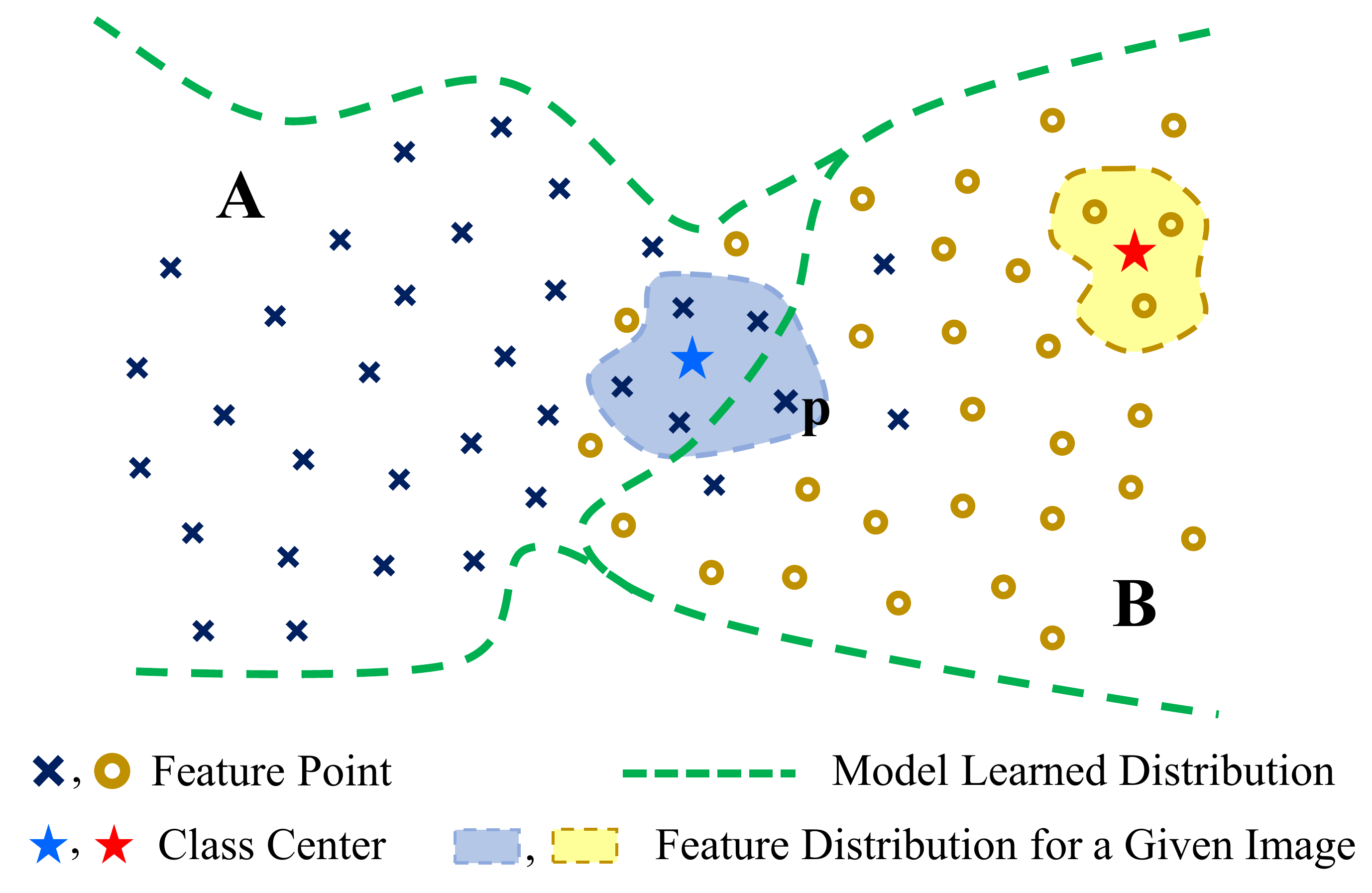}
\end{center}
   \caption{An illustration of the role of class center. For a given pixel $p$ which belongs to class $A$, the model mislabels it to class $B$ when only uses the feature of $p$. But if the model knows about the representation (class center) of $A$ (light blue area) and $B$ (light yellow area) in the image, it can find that $p$ more likely comes from $A$ rather than $B$. Thus, the wrong prediction could be corrected.}
\label{fig:insight}
\end{figure}

\label{section:class_center}
The benefits of class center are two-fold. Firstly, it allows the pixels to understand the overall presentation of each class from a global view. Since the class center is the combination of all pixels in an image, this gives a strong supervision information while training and can help the model learn more discriminative features for each class. Moreover, the class center can also help to check for the consistency between one pixel and each class center in the image to improve the performance. Therefore, the distribution of each class can be further refined. It is known that a model always learns the distribution of each category across the entire dataset, thus for a specific image, the distribution of a particular category often occupies a small portion of the distribution of that category over the entire dataset. So the class center of this portion is more representative and helpful for the pixel classification in this image. By introducing the class center, the model can correct many cases which are wrongly classified before. An example is shown in Figure \ref{fig:insight}, when only the feature of pixel $p$ is used, the model mislabels it to class $B$. But the misclassification can be further fixed by considering the class centers at the same time.

\subsubsection{Attentional Class Feature}

Inspired by the attention mechanism, we present the attentional class feature. Different pixels need to selectively attend to different classes. For a pixel $p$, we use the coarse segmentation result as its attention map to calculate its attentional class feature. The reason why we use the coarse segmentation result is straightforward. If the coarse segmentation mislabels a pixel to a wrong class, it needs to pay more attention to that wrong class to check for the feature consistency. Or if some classes do not even exist in the image, the pixel does not need to know about these classes. As in Figure \ref{fig:insight}, the pixel $p$ only needs to be aware of the class centers of $A$ and $B$ rather than other class centers.

We propose a Class Attention Block (CAB) which is shown in Figure \ref{fig:ccb_acfb} (b) to calculate the attentional class feature. Given the class centers $F_{class} \in \mathbb{R}^{N \times C'}$ and coarse segmentation result $P_{coarse} \in \mathbb{R}^{N \times H \times W}$, we first reshape $P_{coarse}$ to $\mathbb{R}^{N \times HW}$. And then a matrix multiplication is applied to the transpose of $F_{class}$ and $P_{coarse}$ to calculate the attentional class feature $F_a$ for each pixel. More specifically, the attentional class feature of pixel $j$, denoted as $F_a^j$, can be calculated as follows,
\begin{equation}
    F_{a}^j = \sum_{i=0}^N P_{coarse}^{i, j} \cdot F_{class}^i,
\label{equa:attentional}
\end{equation}

where both $F_a^j$ and $F_{class}^i$ are in $\mathbb{R}^{1 \times C'}$. 

After the attentional class feature is calculated, we apply a $1 \times 1$ \emph{conv} to refine the calculated feature.

\subsection{Attentional Class Feature Network}

Based on Attentional Class Feature (ACF) module, we propose the Attentional Class Feature Network for semantic segmentation as illustrated in Figure \ref{fig:overall}. ACFNet consists of two separate parts, base network and ACF module. The base network is a complete segmentation network. In our experiments, we use the ResNet \cite{he2016deep} and ResNet with atrous spatial pyramid pooling (ASPP) \cite{chen2017rethinking} as our base networks respectively to verify the effectiveness of our ACF module. The ACF module leverages the segmentation result and feature map in base network to calculate the attentional class feature. Finally, we concatenate the attentional class feature and the feature map in base network together and refine it through a $1 \times 1$ \emph{conv} to get the final segmentation result.

\textbf{Loss Function.} For explicit feature refinement, we use the auxiliary supervision to improve the performance and make the network easier to optimize following PSPNet \cite{zhao2017pyramid}. The class-balanced cross entropy loss is employed for auxiliary supervision, coarse segmentation and fine segmentation. Finally, we use three parameters $\lambda_a$, $\lambda_c$ and $\lambda_f$ to balance the auxiliary loss $l_a$, the coarse segmentation loss $l_c$ and the fine segmentation loss $l_f$ as shown in Equation. \ref{equa:loss} .
\begin{equation}
    L = \lambda_a \cdot l_a + \lambda_c \cdot l_c + \lambda_f \cdot l_f.
\label{equa:loss}
\end{equation}

\section{Experiments}

To evaluate the proposed module, we conduct several experiments on the Cityscapes \cite{Cordts2016Cityscapes} dataset. The Cityscapes dataset is collected for urban scene understanding, which contains 19 classes for scene parsing or semantic segmentation evaluation. It has 5,000 high resolution ($2048 \times 1024$) images, of which 2,975 images for training, 500 images for validation and 1,525 for testing. In our experiments, we use the mean of class-wise Intersection over Union (mIoU) as the evaluation metric.

\subsection{Network Architecture}
We use two base networks to verify the effectiveness and generality of ACF module. One is ResNet-101 which is our baseline network and the other one is ResNet-101 with ASPP. The experiments on the latter network show that our module can also significantly improve the performance when combined with other state-of-the-art modules.

\textbf{Baseline Network.} As for baseline network, we use the ResNet-101 pre-trained on ImageNet \cite{deng2009imagenet}. Following PSPNet \cite{zhao2017pyramid}, the classification layer and last pooling layer are removed and the dilation rate of the convolution layers within the last two blocks are set to 2 and 4 respectively. The output stride of the network is set to 8.

\textbf{Baseline Network with ASPP.} It is known that the atrous spatial pyramid pooling (ASPP) \cite{chen2017rethinking} has achieved great success in segmentation tasks. To verify the generalization ability of the ACF module, we also conduct several experiments based on the ResNet-101 (baseline network) followed by ASPP module. The ASPP consists of four parallel parts: a $1 \times 1$ convolution branch and three $3 \times 3$ convolution branches with dilation rate being 12, 24 and 36 respectively. In our re-implementation of ASPP module, we follow the original paper but change the output channel from 256 to 512 in all of four branches.

\textbf{Attentional Class Feature Module.} To reduce the computation and the memory usage, we first reduce the channel of input feature of ACF module to 512. The channel number of final output of the ACF module is also set to 512.

\subsection{Implementation Details}
For training, we use the stochastic gradient descent (SGD) optimizer \cite{robbins1951stochastic} with the initial learning rate 0.01, weight decay 0.0005 and momentum 0.9 for Cityscapes dataset. Following the previous works \cite{chen2017rethinking, zhao2017pyramid}, we also employ the `poly' learning rate policy, where the learning rate of current iteration is multiplied by the factor $(1-\frac{iter}{max\_iter})^{0.9}$. The loss weights $\lambda_a$, $\lambda_c$ and $\lambda_f$ in Equation. \ref{equa:loss} are set to 0.4, 0.6 and 0.7 respectively. All experiments are trained on 4$\times$ Nvidia P40 GPUs for 40k iterations with batch size 8.

All BatchNorm layers in our network are replaced by InPlaceABN-Sync \cite{rota2018place}. To avoid overfitting, we also employ the common data augmentation strategies, including random horizontal flipping, random scaling in the range of [0.5, 2.0] and random cropping of $769 \times 769$ image patches following \cite{zhao2017pyramid, yang2018denseaspp}.

\subsection{Ablation Study}

In this subsection, we conduct a series of experiments based on the baseline network to reveal the effect of each component in our proposed module. 

\subsubsection{Attentional Class Feature module}
We first use the atrous ResNet-101 as the baseline network and the final results are obtained by directly upsampling the output. For starters, we evaluate the performance of the baseline network, as shown in Table \ref{tab:ablation}. It should be noted that all our experiments use the \emph{auxiliary supervision}.

\textbf{Ablation for Class Center.} To verify the effect of class center, we first remove the Class Attention Block (CAB) in Figure \ref{fig:overall} (d). The calculated class center $F_{class}$ is reshaped and upsampled to $\mathbb{R}^{NC \times H \times W}$. Then the upsampled class center and the feature map in base network are concatenated to get the fine segmentation result. The experiment result is also shown in Table \ref{tab:ablation}. This modification improves the performance to 76.42\%(0.57\%$\uparrow$) on coarse segmentation and 77.94\% (2.09\%$\uparrow$) on fine segmentation.

\textbf{Ablation for Attentional Class Feature.} We further evaluate the role of attentional class feature. Essentially, the calculation process described in Equation.\ref{equa:attentional} is the weighted summation of class centers in which the weight is coarse segmentation probabilities of each pixel. So we call this approach of calculating the attentional class feature as \textbf{ACF(sum)}. Besides \textbf{ACF(sum)}, we also try another way, named \textbf{ACF(concat)}, to leverage the coarse segmentation probabilities and class centers to get another type of attentional class feature. For a given pixel $j$, \textbf{ACF(concat)} can be formulated as follows,

\begin{equation}
    F_{a}^j = {\rm CONCAT}_{i=0}^{N}\{ P_{coarse}^{i, j} \cdot F_{class}^i\},
\label{equa:attentional_concat}
\end{equation}

where $F_{a}^j$ is in $\mathbb{R}^{NC' \times 1}$ and it is the weighted concatenation of class centers in which the weight is coarse segmentation probabilities of each pixel. The experiment results are shown in Table \ref{tab:ablation}. Compared with the experiment of class center, the  \textbf{ACF(concat)} improves the performance of fine segmentation from 77.94\% to 79.17\% while the \textbf{ACF(sum)} achieves performance of 79.32\%. When comparing with the baseline, the improvement is significant. In the following experiments, we use the \textbf{ACF(sum)} strategy as default.

\begin{table}
\begin{center}
\begin{tabular} {lcc}
\hline
Method & mIoU(\%) \\
\hline
ResNet-101 Baseline & 75.85 \\
\hline
ResNet-101 + class center & 76.42(\textbf{C}) / 77.94(\textbf{F}) \\
ResNet-101 + ACF (concat) & 76.66(\textbf{C}) / 79.17(\textbf{F}) \\
ResNet-101 + ACF (sum) & 76.56(\textbf{C}) / 79.32(\textbf{F}) \\
\hline
\end{tabular}
\end{center}
\caption{Detailed performance comparison of our proposed Attentional Class feature module on Cityscapes val. set based on the ResNet-101. \textbf{C}: result of coarse segmentation. \textbf{F}: result of fine segmentation. \textbf{ACF(concat)}: the attentional class feature is calculated by the weighted concatenation of class centers. \textbf{ACF(sum)}: the attentional class feature is calculated by the weighted summation of class centers.}
\label{tab:ablation}
\end{table}

\subsubsection{Feature Similarity}

\begin{figure}[t]
\begin{center}
   \includegraphics[width=1\linewidth]{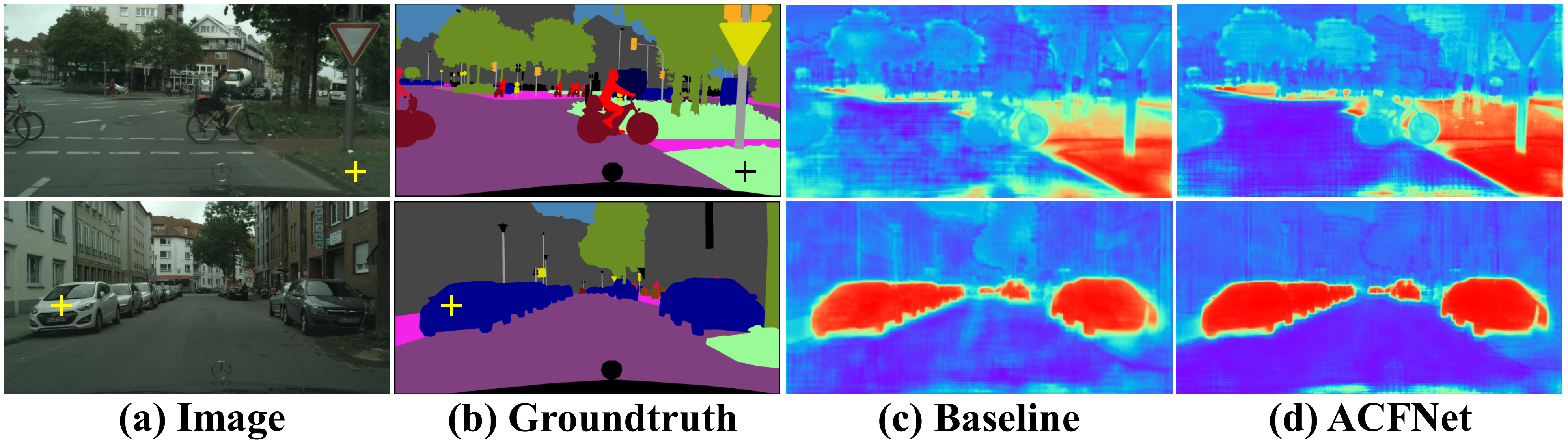}
\end{center}
   \caption{Feature similarity visualization of all pixels to a given pixel. Hotter color denotes more similar in feature level. The pixels we selected are marked as \textbf{cross sign} in (a) Image and (b) Groundtruth. Column (c) and (d) show the similarity maps of all other pixels to the selected pixel of baseline network and ACFNet.}
\label{fig:feature}
\end{figure}

\textbf{Improvement Compared with Baseline.} In order to better understand how ACF module improves the final result, we visualize the cosine similarity map between a given pixel and other pixels in the feature map. As shown in Figure \ref{fig:feature}, we select two pixels from `terrain' and `car' respectively. The feature similarity maps of the baseline and ACFNet are shown in column (c) and (d) separately. For ACFNet, we use the feature map before fine segmentation to calculate the feature similarity. After adding the class-level context, ACFNet learns a more discriminative feature for each class. The intra-class features are more consistent and the inter-class features are more distinguishable. 

\begin{figure}[t]
\begin{center}
   \includegraphics[width=1\linewidth]{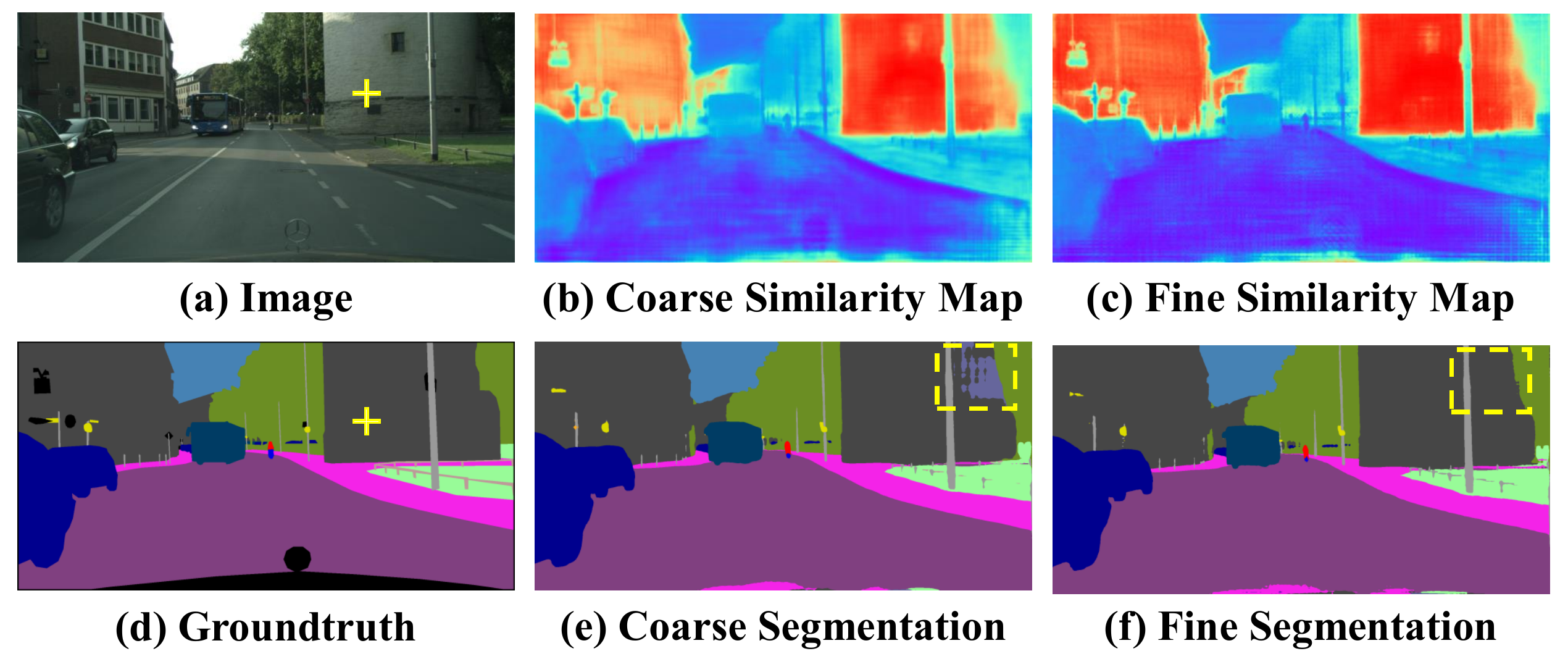}
\end{center}
   \caption{Feature similarity visualization of the feature maps before coarse segmentation and fine segmentation. The pixel selected to calculate the similarity with other pixels is marked by \textbf{cross sign} in (a) and (d). (b) and (c) show the similarity maps of feature maps before coarse segmentation and fine segmentation respectively. And the visual improvement part is marked by yellow square in (e) and (f).}
\label{fig:coarse_to_fine}
\end{figure}

\textbf{Improvement Compared with Coarse Segmentation.} As discussed in section \ref{section:class_center}, the class-level context may also help a pixel check for the consistency with each class in the image and further refine the segmentation result. To verify this idea, we also visualize the feature similarity of the feature maps before coarse segmentation and fine segmentation given a specific pixel. As shown in Figure \ref{fig:coarse_to_fine}, the area which shows the improvement is marked by yellow square in both (e) coarse segmentation and (f) fine segmentation. From (b) and (e), we can see that the model does not learn a good enough distribution of class `building' and thus mislabels a lot of pixels. Features of those mislabeled pixels are inconsistent with those correctly labeled pixels. But after adding the attentional class feature for those pixels, the refined feature shows the consistency between mislabeled pixels and correctly labeled pixels. Thus, the final result has a significant improvement.

\subsubsection{Result Visualization}
\begin{figure*}[t]
\begin{center}
   \includegraphics[width=1\linewidth]{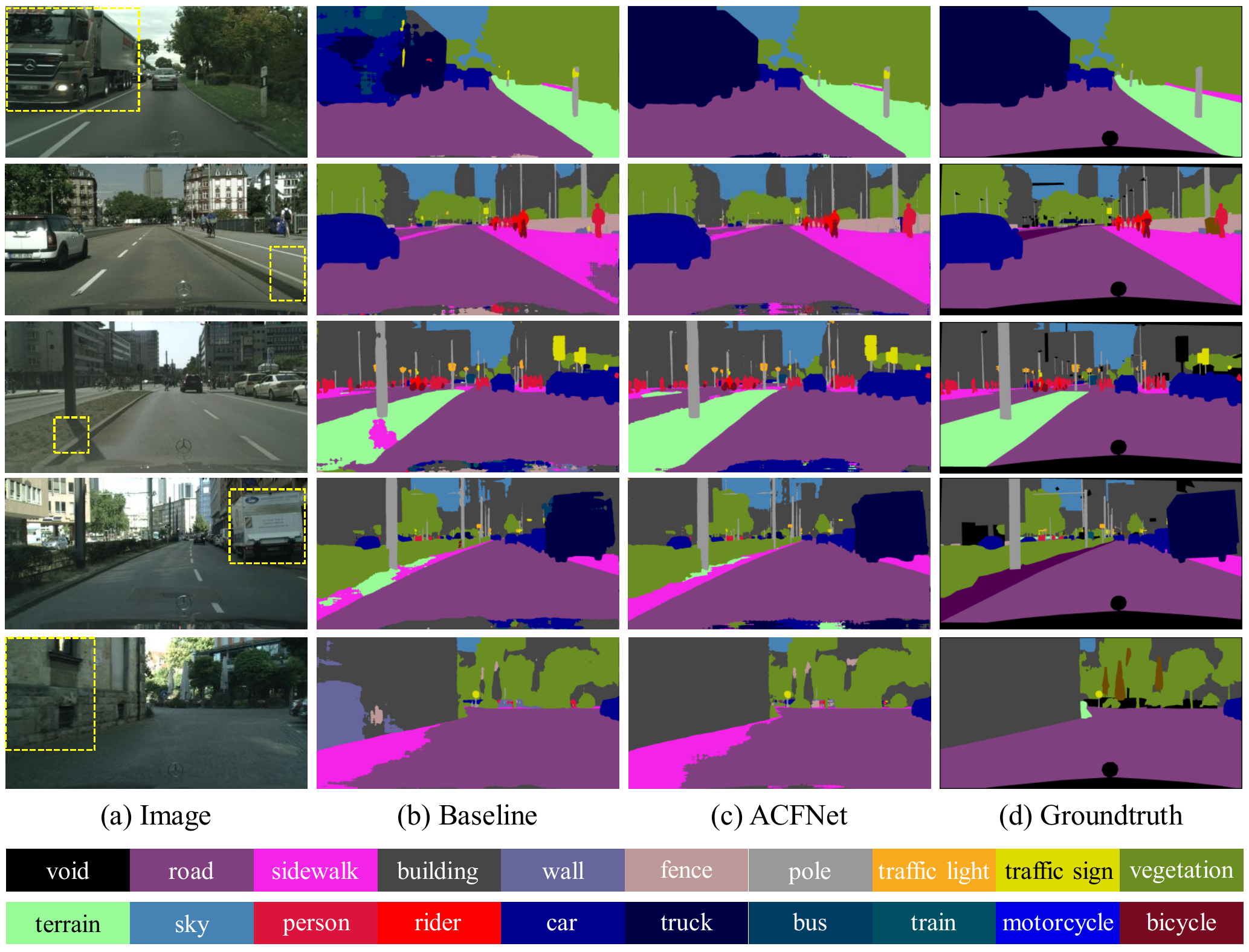}
\end{center}
   \caption{Visualization results of ACFNet based on ResNet-101 network on Cityscapes val set.}
\label{fig:visualize}
\end{figure*}

We provide the qualitative comparisons between ACFNet and baseline network in Figure \ref{fig:visualize}. We use the \emph{yellow square} to mark those challenging regions. The baseline easily mislabels such areas, but ACFNet is able to correct them. For example, the baseline model can not classify `truck' or `car' correctly in the first example and mislabels the `building' and `wall' in the fifth example. After adding the ACF module, such areas are greatly corrected.

\subsection{Experiments on Baseline Network with ASPP}

To verify the generality of ACF module, we also combine it with ResNet-101 and ASPP. We first conduct the baseline (ResNet-101 with ASPP) experiment and the result is shown in Table \ref{tab:ablation_strong}. Our re-implemented version of ASPP achieves similar performance compared with the original paper \cite{chen2017rethinking} (78.42\% \emph{vs.} 77.82\%).

\textbf{Performance with ACF Module.} We append the ACF module to the end of ASPP and the experiment result is shown in Table \ref{tab:ablation_strong}. After adding the ACF module, the performance is improved by 1.7\% (78.42\% to 80.08\%), which verifies that our ACF module can work together with other state-of-the-art modules to further boost the performance.

Moreover, we apply the online bootstrapping \cite{wu2016high} and multi-scale (MS), left-right flipping (Flip) to improve the performance based on the ResNet-101+ ASPP + ACF. The results on Cityscapes val are shown in Table \ref{tab:ablation_strong}.

\begin{compactitem}
    \item \textbf{Online Bootstrapping:} Following the previous works \cite{wu2016high}, we adopt the online bootstrapping for hard training pixels. The hard training pixels are those whose probabilities on the correct classes are less than a certain threshold $\theta$. When training with online bootstrapping, we keep at least $K$ pixels within each batch. In our experiments, we set $\theta$ to 0.7 and $K$ to 100,000. With online bootstrapping, the performance on Cityscapes val set can be improved by 0.91\%.
    \item \textbf{MS/Flip:} As many of previous works \cite{zhao2017pyramid, yu2018learning, fu2018dual, yang2018denseaspp, chen2018encoder}, we also adopt the left-right flipping and multi-scale $[0.75 , 1.0 , 1.25 , 1.5 , 1.75 , 2.0]$ strategies while testing. From Table \ref{tab:ablation_strong}, we can see that \textbf{MS/Flip} improves the performance by 1.38\% on val set.
\end{compactitem}

\begin{table}
\begin{center}
\begin{tabular} {lcc}
\hline
Method & mIoU(\%) \\
\hline
ResNet-101 + ASPP Baseline & 78.42 \\
\hline
ResNet-101 + ASPP + ACF & 80.08 \\
ResNet-101 + ASPP + ACF + OB & 80.99 \\
ResNet-101 + ASPP + ACF + MS/Flip & 81.46 \\
\hline
\end{tabular}
\end{center}
\caption{Detailed performance comparison of our proposed Attentional Class feature module on Cityscapes val. set base on the ResNet-101 with ASPP. \textbf{ACF }: attentional class feature module. \textbf{OB}: using online bootstrapping while training. \textbf{MS/Flip}: using multi-scale and flipping while testing.}
\label{tab:ablation_strong}
\end{table}

\subsection{Comparing with the State-of-the-Art}

We further compare ACFNet with the existing methods on the Cityscapes test set by submitting our result to the official evaluation server. Specifically, we train the ResNet-101 with ASPP and ACF with online bootstrapping strategy and use the multi-scale \& flipping strategies while testing. The results and comparison are illustrated in Table \ref{tab:sota}. ACFNet, which uses only \emph{train-fine} data, outperforms previous work PSANet \cite{zhao2018psanet} for about \textbf{2.2\%} and even better than most methods that also employ the validation set for training. While using both \emph{train-fine} and \emph{val-fine} data for training, ACFNet outperforms the previous methods \cite{yang2018denseaspp, zhao2018psanet, yu2018learning, yu2018bisenet} for a large margin and achieves new state-of-the-art of 81.85\% mIoU.

\begin{table*}
\scalebox{0.73}{
\begin{tabular} {l|c|cccccccccccccccccccc}
Methods & \rotatebox{90}{Mean IoU} & \rotatebox{90}{road} & \rotatebox{90}{sidewalk} & \rotatebox{90}{building} & \rotatebox{90}{wall} & \rotatebox{90}{fence} & \rotatebox{90}{pole} & \rotatebox{90}{traffic light} & \rotatebox{90}{traffic sign} & \rotatebox{90}{vegetation} &  \rotatebox{90}{terrain} & \rotatebox{90}{sky} & \rotatebox{90}{person} & \rotatebox{90}{rider} & \rotatebox{90}{car} & \rotatebox{90}{truck} & \rotatebox{90}{bus} & \rotatebox{90}{train} & \rotatebox{90}{ motorcycle} & \rotatebox{90}{bicycle} \\
\hline
\hline
PSPNet \dag \cite{zhao2017pyramid} & 78.4 & - & - & - & - & - & - & - & - & - & - & - & - & - & - & - & - & - & - & -  \\
PSANet \dag \cite{zhao2018psanet} & 78.6 & - & - & - & - & - & - & - & - & - & - & - & - & - & - & - & - & - & - & -  \\ 
ACFNet (ours) \dag  & \textbf{80.8} & 98.7 & 87.1 & 93.7 & 60.8 & 62.0 & 69.7 & 77.7 & 80.4 & 94.0 & 73.6 & 95.7 & 87.6 & 73.6 & 96.1 & 65.6 & 87.3 & 83.0 & 70.5 & 78.0 \\
\hline
\hline
DeepLab-v2 \cite{chen2018deeplab} & 70.4 & 97.9 & 81.3 & 90.3 & 48.8 & 47.4 & 49.6 & 57.9 & 67.3 & 91.9 & 69.4 & 94.2 & 79.8 & 59.8 & 93.7 & 56.5 & 67.5 & 57.5 & 57.7 & 68.8 \\
RefineNet \ddag \cite{lin2017refinenet} & 73.6 & 98.2 & 83.3 & 91.3 & 47.8 & 50.4 & 56.1 & 66.9 & 71.3 & 92.3 & 70.3 & 94.8 & 80.9 & 63.3 & 94.5 & 64.6 & 76.1 & 64.3 & 62.2 & 70 \\
GCN \ddag \cite{peng2017large} & 76.9 & - & - & - & - & - & - & - & - & - & - & - & - & - & - & - & - & - & - & -  \\ 
DUC \ddag \cite{wang2018understanding} & 77.6 & 98.5 & 85.5 & 92.8 & 58.6 & 55.5 & 65 & 73.5 & 77.9 & 93.3 & 72 & 95.2 & 84.8 & 68.5 & 95.4 & 70.9 & 78.8 & 68.7 & 65.9 & 73.8 \\  
ResNet-38 \cite{wu2019wider} & 78.4 & 98.5 & 85.7 & 93.1 & 55.5 & 59.1 & 67.1 & 74.8 & 78.7 & 93.7 & 72.6 & 95.5 & 86.6 & 69.2 & 95.7 & 64.5 & 78.8 & 74.1 & 69 & 76.7 \\
BiSeNet \ddag \cite{yu2018bisenet} & 78.9 & - & - & - & - & - & - & - & - & - & - & - & - & - & - & - & - & - & - & -  \\ 
DFN \ddag \cite{yu2018learning} & 79.3 & - & - & - & - & - & - & - & - & - & - & - & - & - & - & - & - & - & - & -  \\ 
PSANet \ddag \cite{zhao2018psanet} & 80.1 & - & - & - & - & - & - & - & - & - & - & - & - & - & - & - & - & - & - & -  \\ 
DenseASPP \ddag \cite{yang2018denseaspp} & 80.6 & \textbf{98.7} & \textbf{87.1} & 93.4 & \textbf{60.7} & 62.7 & 65.6 & 74.6 & 78.5 & 93.6 & 72.5 & 95.4 & 86.2 & 71.9 & 96.0 & \textbf{78.0} & \textbf{90.3} & 80.7 & 69.7 & 76.8 \\ 
ACFNet (ours) \ddag & \textbf{81.8} & \textbf{98.7} & \textbf{87.1} & \textbf{93.9} & 60.2 & \textbf{63.9} & \textbf{71.1} & \textbf{78.6} & \textbf{81.5} & \textbf{94.0} & \textbf{72.9} & \textbf{95.9} & \textbf{88.1} & \textbf{74.1} & \textbf{96.5} & 76.6 & 89.3 & \textbf{81.5} & \textbf{72.1} & \textbf{79.2}  \\
\hline
\end{tabular}}
\begin{tablenotes} 
        \item \footnotesize{\dag Training with only the \emph{train-fine} dataset.}
        \item \footnotesize{\ddag Training with both the \emph{train-fine} and \emph{val-fine} datasets.}
\end{tablenotes}
\caption{Per-class results on  Cityscapes test set with the state-of-the-art models. ACFNet outperforms existing methods and achieves 81.8\% in mIoU.}
\label{tab:sota}
\end{table*}

\section{Conclusion}

In this paper, we propose the concept of \emph{class center} to represent the class-level context to improve the segmentation performance. We further propose a coarse-to-fine segmentation structure based on our attentional class feature module, called ACFNet, to calculate and selectively combine the class-level context according to the feature of each pixel. The ablation studies and visualization of intermediate results show the effectiveness of class-level context. ACFNet achieves new state-of-the-art on Cityscapes dataset with mIoU of 81.85\%.

\section{Acknowledgment}

Feifei Ma is supported by the Youth Innovation Promotion Association, Chinese Academy of Sciences. Besides, our special thanks go to Yuchen Sun, Xueyu Song, Ru Zhang, Yuhui Yuan and the anonymous reviewers for the discussion and their helpful advice. 

{\small
\bibliographystyle{ieee_fullname}
\bibliography{egbib}
}

\end{document}